\documentclass[acmtog, nonacm]{acmart}
\renewcommand{\setcopyright}[1]{}

\usepackage{booktabs} 

\usepackage[ruled]{algorithm2e} 

\SetAlFnt{\small}
\SetAlCapFnt{\small}
\SetAlCapNameFnt{\small}
\SetAlCapHSkip{0pt}

\acmJournal{TOG}

\newcommand{\boldparagraph}[1]{\vspace{0.1cm}\noindent{\bf #1.}}
\newcommand{\tightboldparagraph}[1]{\noindent{\bf #1.}}

\newcommand{\Resolution}{\mathrm{R}}
\newcommand{\Opacity}{\alpha}
\newcommand{\Color}{\mathbf{c}}

\newcommand{\SurfaceMultiplier}{s}
\newcommand{\SurfaceCounter}{\mathcal{S}}
\newcommand{\FreespaceCounter}{\mathcal{F}}
\newcommand{\ObservedCounter}{\mathcal{O}}
\newcommand{\FreespaceThreshold}{t}
\newcommand{\VoxelDistance}{D}

\newcommand{\DataBlockSize}{\mathrm{D}}

\newcommand\rgt{\aftergroup\mathclose\aftergroup{\aftergroup}\right}

\usepackage{colortbl}
\usepackage{makecell}
\usepackage{soul}
\usepackage{enumitem}
\definecolor{yellow}{rgb}{1,1, 0.7}
\definecolor{lightyellow}{rgb}{1,1, 0.8}
\definecolor{orange}{rgb}{1, 0.85, 0.7}
\definecolor{tablered}{rgb}{1, 0.7, 0.7}
\usepackage[capitalize]{cleveref}
\crefname{section}{Sec.}{Secs.}
\Crefname{section}{Section}{Sections}
\Crefname{table}{Table}{Tables}
\crefname{table}{Tab.}{Tabs.}

\usepackage{multirow}

\definecolor{RowColorA}{rgb}{1.,0.94,0.94}
\definecolor{RowColorB}{rgb}{0.94,0.94,1}
\definecolor{RowColorAIntense}{rgb}{0.6,0.25,0.25}
\definecolor{RowColorBIntense}{rgb}{0.25,0.25,0.6}

\newcommand{\norm}[1]{\left| \left| #1 \right| \right|}

\newcommand{\SurfaceDepth}{d}

\begin{document}
\title{Binary Opacity Grids: Capturing Fine Geometric Detail for Mesh-Based View Synthesis}

\author{Christian Reiser}
\affiliation{
\institution{University of Tübingen, Tübingen AI Center, Google Research}
\country{Germany}
\city{Tübingen}
}
\email{christian.j.reiser@gmail.com}
\orcid{0009-0002-1050-3958}

\author{Stephan Garbin}
\affiliation{
\institution{Google Research}
\country{United Kingdom}
\city{London}
}
\email{stephangarbin@outlook.com}
\orcid{TODO}

\author{Pratul P. Srinivasan}
\affiliation{%
\institution{Google Research}
\country{United States of America}
\city{San Francisco}
}
\email{pratuls@google.com}
\orcid{0009-0008-8268-3285}

\author{Dor Verbin}
\affiliation{%
\institution{Google Research}
\country{United States of America}
\city{Boston}
}
\email{dorverbin@google.com}
\orcid{0000-0001-8798-3270}

\author{Richard Szeliski}
\affiliation{
\institution{Google Research}
\country{United States of America}
\city{Seattle} 
}
\email{szeliski@google.com}
\orcid{0009-0005-5300-5475}

\author{Ben Mildenhall}
\affiliation{%
\institution{Google Research}
\country{United States of America}
\city{San Francisco}
}
\email{bmild@google.com}
\orcid{0009-0001-9796-6121}

\author{Jonathan T. Barron}
\affiliation{%
\institution{Google Research}
\country{United States of America}
\city{San Francisco}
}
\email{barron@google.com}
\orcid{0009-0008-4016-9448}

\author{Peter Hedman}
\authornote{Equal advising.}
\affiliation{%
\institution{Google Research}
\country{United Kingdom}
\city{London}
}
\email{hedman@google.com}
\orcid{0000-0002-2182-0185}

\author{Andreas Geiger}
\authornotemark[1]
\affiliation{%
\institution{University of Tübingen, Tübingen AI Center}
\country{Germany}
\city{Tübingen}
}
\email{a.geiger@uni-tuebingen.de}
\orcid{0000-0002-8151-3726}

\renewcommand{\shortauthors}{Reiser, Garbin, Srinivasan, Verbin, Szeliski, Mildenhall, Barron, Hedman, Geiger}

\authorsaddresses{}

\begin{abstract}
While surface-based view synthesis algorithms are appealing due to their low computational requirements, they often struggle to reproduce thin structures. In contrast, more expensive methods that model the scene’s geometry as a volumetric density field (e.g. NeRF) excel at reconstructing fine geometric detail. However, density fields often represent geometry in a ``fuzzy'' manner, which hinders exact localization of the surface. In this work, we modify density fields to encourage them to converge towards surfaces, without compromising their ability to reconstruct thin structures. First, we employ a discrete opacity grid representation instead of a continuous density field, which allows opacity values to discontinuously transition from zero to one at the surface. Second, we anti-alias by casting multiple rays per pixel, which allows occlusion boundaries and subpixel structures to be modelled without using semi-transparent voxels. Third, we minimize the binary entropy of the opacity values, which facilitates the extraction of surface geometry by encouraging opacity values to binarize towards the end of training. Lastly, we develop a fusion-based meshing strategy followed by mesh simplification and appearance model fitting.
The compact meshes produced by our model can be rendered in real-time on mobile devices and achieve significantly higher view synthesis quality compared to existing mesh-based approaches.

\end{abstract}

\begin{teaserfigure}
\vspace{-0.3cm}
\begin{flushleft}
  {\LARGE Interactive webdemo at \textcolor{blue}{\texttt{\href{https://binary-opacity-grid.github.io}{https://binary-opacity-grid.github.io}}}}\\
  \vspace{0.2cm}
  \includegraphics[width=\linewidth]{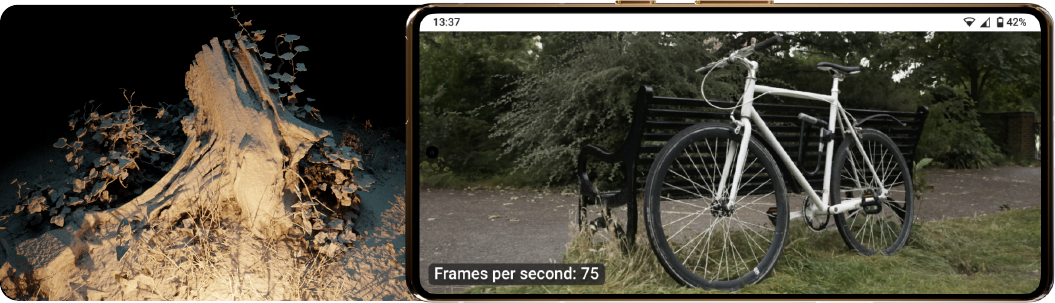}
  \vspace{-0.25in}
  \caption{Our method reconstructs triangle meshes from multi-view images and is able to capture fine geometric detail such as leaves, branches and grass (left). At the same time our meshes are compact enough for real-time view synthesis on a Google Pixel 8 Pro (right).}
  \label{fig:teaser}
\end{flushleft}
\end{teaserfigure}



\maketitle


\section{Introduction}

\begin{figure}
  \centering
  \includegraphics[width=\linewidth]{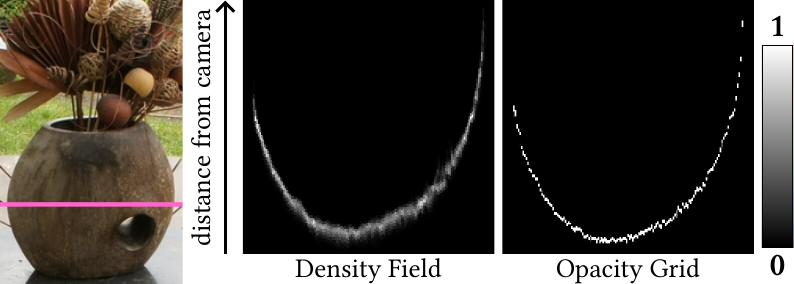}
  \vspace{-0.25in}
  \caption{We visualize the volume rendering weights for rays corresponding to a row of pixels (left, in pink). Density fields such as Zip-NeRF~\cite{Zip-NeRF} tend to represent hard surfaces as semi-transparent volumes despite their use of surface-promoting regularizers. In contrast, our opacity grid converges to a hard surface.
  Note that, since each pixel column visualizes the volume rendering weights of a single ray, gaps in this visualization do not indicate the presence of holes in the underlying representation.
  }
  \label{fig:vol_render_weight_viz}
\end{figure}

Surface rendering is generally considered to be more efficient than volume rendering, as surface rendering ideally only requires reading appearance data from a single 3D location, while volume rendering requires aggregating colors and densities across multiple points along each ray. Nevertheless, the current highest-quality view synthesis algorithms~\cite{3DGS, Zip-NeRF, SMERF} all use volume rendering. These algorithms tend to represent even hard surfaces as ``fuzzy'' volumes, which leads to their high computational cost. This also holds when applying surface-promoting regularizers~\cite{Mip-NeRF_360}, see Figure~\ref{fig:vol_render_weight_viz}.


Recently, BakedSDF~\cite{BakedSDF} has demonstrated that accurate view synthesis is also possible with a surface-based approach. However, in contrast to volumetric methods, BakedSDF struggles with recovering fine geometric detail. One reason for this is that BakedSDF adopts the currently dominant paradigm for 3D reconstruction, where the SDF is converted to a fuzzy volume during training~\cite{VolSDF, NeuS, NeuralAngelo, MonoSDF}. This soft conversion from SDF to volumetric density allows the model to 
``cheat'' by representing thin structures in a fuzzy manner. As a result, during meshing, thin structures often vanish. Furthermore, during training, the validity of the recovered SDF must be ensured using an Eikonal loss, which acts as a smoothness prior and thereby tends to remove fine geometric detail.



To avoid these weaknesses of SDF-based approches, we investigate an alternative strategy that does not require an Eikonal loss or soft density conversion. We use a volume-based representation whose geometry we successively ``sharpen'' during training. We achieve this surface convergence by applying the following three modifications to an existing state-of-the-art radiance field model~\cite{Zip-NeRF}. First, we employ a discrete opacity grid instead of a continuous density field, which enables opacity values to discontinuously transition from zero to one at the surface~\cite{MobileNeRF}. Second, we cast multiple rays per pixel to allow our model to accurately reproduce anti-aliased occlusion boundaries without using semi-transparent voxels. Third, we explicitly encourage hard surfaces by enforcing a binary entropy loss on the opacity values. As shown in Figure~\ref{fig:vol_render_weight_viz}, this causes opacity values to binarize to zero or one as training converges, which enables the extraction of surface geometry. We demonstrate that all three of these elements are required for accurate reconstruction of subpixel structures.

Furthermore, we present a fusion-based meshing strategy for converting our recovered binary opacity grid into a triangle mesh after training. The resulting mesh can then be simplified with off-the-shelf tools to a complexity that is adequate for real-time rendering while still preserving thin structures. Finally, we equip that mesh with a lightweight view-dependent appearance model that is well-suited for real-time viewer applications. Because the standard approach of UV mapping is problematic for our highly detailed meshes, we systematically evaluate alternative appearance representations and find the combination of triplanes with a low-resolution voxel grid as the preferred method~\cite{MERF}. Our triangle mesh and appearance representation are compact enough to be rendered in real-time on mobile devices and achieve significantly higher view synthesis quality compared to existing mesh-based models. As such, our work represents a step towards closing the gap between surface-based view synthesis methods and volume-based ones.

\section{Related Work}
In this section, we review real-time view synthesis methods that fit a 3D scene representation to calibrated multi-view images using differentiable rendering. These methods can be subdivided according to their rendering formulation into volume-based, surface-based and hybrid methods.

\boldparagraph{Volume-based Methods} The highest-quality view synthesis methods use volume rendering during optimization and inference \cite{Zip-NeRF}. 
Many techniques in this category follow the neural radiance fields (NeRFs) paradigm \cite{NeRF}. NeRFs associate a density and view-dependent color value with each 3D point and can be fitted using differentiable volume rendering to multi-view images with a re-rendering objective. The original NeRF uses an MLP to represent the scene, which results in slow rendering. Follow-up works speed-up rendering by using alternative representations such as voxel grids~\cite{SNeRG, PlenOctree, FastNeRF, PlenVDB}, triplanes~\cite{TensoRF, MERF, SMERF}, or point-based representations~\cite{Neat, PointNeRF, ADOP, PBNR, 3DGS}.

\boldparagraph{Surface-based Methods} Although recent volume-based methods such as SMERF or 3DGS are capable of real-time rendering, they are slower than surface-based alternatives such as BakedSDF, as demonstrated in \citet{SMERF}. In volume rendering, computing a pixel's value requires compositing colors from \emph{multiple} sampling locations (SMERF) or primitives (3DGS), while surface rendering (under typical conditions) only requires reading appearance data from a \emph{single} surface location. This is the case for surface-based view synthesis methods that employ a triangle mesh, which can be efficiently rendered using hardware-accelerated rasterization.

Early view-synthesis methods used surface geometry from multi-view stereo~\cite{schoenberger2016mvs,jancosek2011multi} and modelled appearance by blending between input images~\cite{debevec1998efficient,wood2000surface,waechter2014let}.
Later methods improved visual quality by predicting the appearance of the surface with a trained neural network~\cite{thies2019neural,riegler2021svs,philip2021free}.
However, these approaches were constrained by the quality of the reconstructed surface, as they did not jointly optimize for appearance and geometry.

MobileNeRF is similar to our method in that it also uses an opacity-based representation during training~\cite{MobileNeRF}. The key difference is that MobileNeRF outputs a coarse proxy mesh equipped with binary alpha masks, whereas we aim for a traditional mesh, which enables wider compatibility. To this end, we use a significantly higher voxel grid resolution than MobileNeRF and rely on simplification to obtain a compact mesh. Our fine grid is more geometrically expressive than MobileNeRF's alpha-textured coarse mesh, which is unable to represent multiple close-by sheets of geometry. During training, we achieve surface convergence using a combination of supersampling and entropy regularization, while MobileNeRF differentiably quantizes opacity values.


UNISURF also uses an opacity-based representation~\cite{UNISURF}. In addition to volume rendering, UNISURF uses a second rendering formulation where the $0.5$ level set defines the surface \cite{DVR, IDR}. We instead obtain hard surfaces by regularizing opacity values.

BakedSDF optimizes a signed distance function (SDF) that can be converted into a mesh after training and encodes appearance as vertex attributes~\cite{BakedSDF}. Similar to VolSDF~\cite{VolSDF}, NeuS~\cite{NeuS}, or NeuralAngelo~\cite{NeuralAngelo}, BakedSDF converts signed distances to density values during optimization, and those densities are used for volume rendering. Valid SDFs are encouraged through the use of a loss to enforce the Eikonal constraint, but this constraint is sometimes violated in favor of reconstructing fine geometric detail in a fuzzy manner. As a result, thin structures often vanish when ``baking'' these SDFs into meshes. In contrast, in our method, there is high agreement between optimized and extracted geometry, since opacity values mostly become binary towards the end of the training.

A number of recent papers also focus on fine geometric detail. NeRFMeshing and NeRF2Mesh both convert a density field into a triangle mesh~\cite{NeRFMeshing, NeRF2Mesh}. Since density fields do not have a clearly defined surface, these methods compensate for lossy mesh conversion with an additional optimization stage. LoD-NeuS uses an error-guided SDF growth strategy to featurizes conical frusta along each ray~\cite{LoD-NeuS}.

DMTet and FlexiCubes differentiably convert an implicit representation to a triangle mesh during training. Similar to our method, any mismatch between optimized and baked geometry is avoided, but these methods do not scale to high resolutions because they require that the full grid be processed during each forward pass~\cite{DMC, DMTet, DMTetFollowUp, FlexiCubes}.

\boldparagraph{Hybrid Methods}
Recently, some methods have emerged that use a combination of surface and volume rendering during inference~\cite{HybridNeRF, VMesh, AdaptiveShells}. These models aim to model the majority of the scene as surface geometry, while modeling whatever small subsets of the scene that happen to look ``fuzzy'' as volumes. Our goal is to expand the portion of the scene that can be represented as a surface, aiming to use volume rendering as sparingly as possible to ensure optimal performance.


\section{Binary Opacity Grids}
To capture thin structures with a surface-based representation, our model first uses an opacity-based voxel grid representation. Through the use of an entropy regularizer and supersampling, our opacity values become binary (either zero or one) towards the end of training. This enables us to exactly locate the surface, which is essential for the conversion of our recovered model into a triangle mesh.

\subsection{Representation}
During training, we represent the scene with an $\Resolution\times\Resolution\times\Resolution$ voxel grid, using a 3D contraction function, as described in Appendix C. With each voxel, we associate an opacity value $\Opacity \in [0,1]$ and a 
color value $\Color \in [0, 1]^3$, which also depends on the view direction. To render a pixel, we cast a ray from the camera origin through the center of the pixel, and this ray is then intersected with all of the voxels along its path. For each intersected voxel, we query its opacity value $\Opacity_k$ and its color value $\Color_k$. The final pixel value $\mathbf{C}$ is computed using front-to-back alpha compositing:
\begin{equation}\label{eq:opacity_rendering}
     \mathbf{C} = \sum_{k} \Opacity_k \left(\prod_{j=1}^{k-1} \Opacity_j \right) \Color_k\,.
\end{equation}
Following MobileNeRF, we directly parameterize opacity values in $[0, 1]$, unlike NeRF, which parameterizes density values that are later converted to opacity values using the distance between sampling points~\cite{NeRF}. In contrast to density-based volume rendering, our formulation does not involve any approximation since it is a finite sum over values associated with the voxels along the ray. One advantage of our formulation is that, when all opacity values are binary, the surface must be located at the first voxel along the ray with an opacity value of one~\cite{MobileNeRF}.

To represent thin structures, we require a high voxel grid resolution $\Resolution$ on the order of $2^{13}$. Directly optimizing a voxel grid of this size is not feasible, as this would require $>\!2$ terabytes of memory to store opacity values alone. Instead, we predict the grid values using an MLP equipped with a multi-resolution hash encoding as in~\citet{INGP}. Note that our overall representation is still discrete in nature because the MLP is only queried at quantized positions \cite{MERF}.

The number of voxels that intersect a ray is proportional to the grid resolution $\Resolution$. Therefore, with a high resolution, it becomes computationally intractable to query the representation at all intersected voxels. To address this, prior work adopted a coarse-to-fine strategy in combination with empty space skipping, but it has been observed that this can cause thin structures to be lost during early training iterations~\cite{NSVF, INGP}. For standard density-based NeRFs, this issue can be circumvented with hierarchical sampling using a ``proposal'' MLP~\cite{NeRF, Mip-NeRF_360}. Howeer, this strategy relies on the assumption that, at the beginning of training, the volume rendering integral can be well-approximated  with randomly placed samples due to the initial volume being somewhat smooth. Since opacity-based rendering does not incorporate the distance between sample points, the finite sum in Equation~ \eqref{eq:opacity_rendering} can only be poorly estimated with a small number of randomly placed samples. To circumvent this, we first train a Zip-NeRF to produce a converged proposal MLP, which encodes the coarse geometry of the scene \cite{Zip-NeRF}. When training our model, we query our representation only at a fixed number of samples from the distribution predicted by the pre-trained proposal MLP, whose weights are kept fixed. These samples represent a superset of the actual surface locations, which entails that the finite sum in Equation~\eqref{eq:opacity_rendering} is computed accurately. If more than one sampled position falls within the same voxel, only the first position is used, which ensures that each voxel only contributes at most once.

\subsection{Training strategy}
Locating a surface and extracting a triangle mesh from an opacity grid requires binary opacity values, but optimizing the opacity values of our grid with no additional regularization does not naturally result in binarized values at the end of training. To encourage binary opacity values, we use an entropy loss that pulls opacity values smaller than 0.5 towards 0 and opacity values larger than 0.5 towards 1. We apply this loss to the opacity values $\alpha_k$ of all voxels sampled along each ray:
\begin{equation}
    \mathcal{L}_{\mathrm{ent}} = \frac{1}{k} \sum_k \mathrm{H}(\alpha_k),
\end{equation}
where $\mathrm{H}$ is the binary entropy function:
\begin{equation}
     \mathrm{H}(p) = -p\log_2(p)-(1-p)\log_2(1-p).
\end{equation}

This alone, however, is not sufficient to accurately reconstruct fine geometric detail. This is because, in a properly anti-aliased image (such as the photographs we use as inputs), each pixel's value is the integral of all light within the cone associated with that pixel.

Consider the case of a ``mixed pixel'' at an occlusion boundary, where a pixel's value depends on light emitted from both a foreground object and a background object.
In volumetric methods such as NeRF or 3DGS, such a pixel will be modeling by reconstructing a semi-transparent region of the foreground object, such that the ray being cast partially penetrates it and proceeds to the background object. This correctly yields a reconstructed pixel value that contains contributions from both the foreground and background objects. But this use of semi-transparency violates the binary entropy assumption required by our model: if opacity values are all binary, casting a single ray through the center of a pixel will result in \emph{either} the foreground or the background object being struck, and will therefore yield an incorrect and aliased pixel intensity (i.e., ``jaggies'').
It is therefore infeasible to accurately reconstruct these mixed pixels using binary opacity values, assuming a single ray is cast for each pixel.  To correctly disambiguate the contributions from multiple surfaces, we therefore cast \emph{multiple} rays per pixel during training. More specifically, we uniformly sample $16$ sub-rays within the footprint of each pixel. After rendering each sub-ray, the final pixel value is computed as the arithmetic mean of the subpixel values. We observe that supersampling produces a significant improvement in geometric quality, especially regarding the reconstruction of thin structures, which often cover less than a single pixel.

\section{Mesh Conversion}
After optimization, we convert the recovered binary opacity grid into a triangular mesh --- the most ubiquitous and practical representation for geometry in computer graphics. If done na\"ively, this conversion leads to a mesh consisting of billions of tiny cubes, which is prohibitively large for real-time rendering. To mitigate this, we design a simple and scalable baking pipeline that outputs a mesh that can be simplified using off-the-shelf tools.

\begin{figure}
  \centering
  \includegraphics[width=0.9\linewidth]{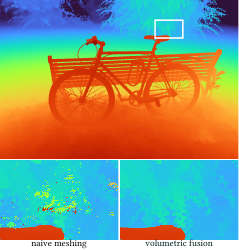}
  \vspace{-0.10in}
  \caption{\textbf{Comparison between different meshing strategies.} The bottom left image shows a depth map rendered from a mesh that was obtained by applying the meshing strategy from \citet{BakedSDF} to our representation. Geometry is instantiated at all visible voxels with an opacity value of $1$ that are sampled by the proposal MLP in \textit{any} training view. This leads to numerous floating artifacts, as infrequently sampled voxels in free space are severely underconstrained by the training loss. The bottom right shows that these underconstrained voxels can be effectively filtered by running volumetric fusion on depth maps rendered from our model. This filtering step also fully preserves thin structures, as can be seen in the top image.}
  \label{fig:counting_based_fusion}
  \vspace{-0.25in}
\end{figure}

\subsection{Volumetric Fusion for Outlier Removal}
The most basic strategy for converting our binary occupancy grid representation into a triangle mesh is to simply instantiate a surface quad between every pair of voxels with opposing opacity values. This works poorly, because the opacity values of voxels in free space are completely unconstrained, as these voxels are never sampled during training. Similarly, occluded space is not constrained by the re-rendering objective, which leads to arbitrary opacity values in the interior of objects. Therefore this strategy results in the creation of many random surfaces, which are either distracting floating artifacts in front of objects or invisible but computationally-wasteful pseudo-geometry in the interior of objects.

A better strategy is to incorporate the proposal MLP that encodes which opacity values are constrained by the training objective. Prior work does this by rendering all training views using the proposal MLP~\cite{MERF, BakedSDF} and then only instantiating surfaces in the vicinity of voxels that contribute to rendering of any pixel. This leads to the filtering of unconstrained areas, since only parts of the scene that are not occluded and sampled by the proposal MLP and thus receive supervision are considered for meshing. However, in our model, this strategy still produces a significant number of floating artifacts, as can be seen in the bottom left image of Figure~\ref{fig:counting_based_fusion}. This is because some voxels are severely underconstrained, as they are not consistently sampled by the proposal MLP during training. In other words, during training,  some voxels are only sampled in a fraction of the training views that observe a voxel. These underconstrained voxels may get erroneously assigned an opacity value of $1$ despite being far from any surface. Since these voxels are still sampled in some of the training views, they are incorrectly appended to the final mesh. To filter these false positives, we employ volumetric fusion~\cite{VolumetricFusion} as described in Appendix A. The bottom right image of Figure~\ref{fig:counting_based_fusion} demonstrates the effectiveness of volumetric fusion for removing outliers.

Another important motivation for using volumetric fusion is that it outputs a dense implicit representation of the scene. As shown in \citet{VolumetricFusion}, this implicit representation can be converted into a hole-free mesh, which is the preferred input for most mesh simplification algorithms. Before conversion to a mesh with marching cubes, we filter the implicit representation with a small Gaussian blur with $\sigma = 1$ to remove geometric noise in the underconstrained outer parts of the scene.

\subsection{Simplification and Visibility Culling}
To produce a more compact representation, we simplify the mesh with an off-the-shelf tool based on quadric edge collapse decimation~\cite{MeshSimplification}. We found this approach to dramatically simplify our meshes while still preserving thin structures. We explicitly simplify the mesh in far-away regions more aggressively, as described in Appendix C. After simplification, we cull triangles that are not visible from any training camera, which leads to another significant reduction in the number of triangles. Only using the training cameras' poses for visibility estimation leads to holes in the mesh that become apparent during novel view synthesis. To combat this, we augment the set of camera poses used for visibility estimation: we create additional poses by adding randomly sampled offsets and rotations to the poses of the training cameras, as described in Appendix C. We find that it is crucial to perform culling \textit{after} simplification, as mesh simplification methods tend to not be robust to the numerous small holes introduced by culling.

\section{View-Dependent Appearance for Meshes}\label{sec:view_dep_app}
To enable view synthesis we need a view-dependent appearance model for our reconstructed mesh.
To this end, we evaluate a number of potential representations and encodings for view-dependent color, with a focus on options that are suited for real-time rendering.

\begin{table}[b]
\centering
\caption{Comparison between representations for mesh appearance on \texttt{gardenvase}. Replacing vertex attributes by a grid-based representation leads to higher quality. However, the sparse voxel grid representation leads to a high memory consumption (VRAM). At a slight loss of quality, the ``triplane + voxel'' option is considerably more compact, while having the fastest rendering among all alternatives.} \vspace{-8pt}
\resizebox{\linewidth}{!}{
\begin{tabular}{@{}l|cccccc@{}}
& PSNR $\uparrow$ & SSIM $\uparrow$ & LPIPS $\downarrow$ & VRAM $\downarrow$ & FPS $\uparrow$ \\ \hline
vertex attributes & 25.58 & 0.771 & 0.211 & \textbf{97} & 261 \\
volume textures & \textbf{26.25} & \textbf{0.820} & \textbf{0.143} & 4513 & 169 \\
triplane + voxel & 26.02 & 0.807 & 0.157 & 629 & \textbf{477} \\ \hline
offline & 26.86 & 0.830 & 0.135 & -- & --
\end{tabular}
}
\label{tab:appearance_representations}
\end{table}

\subsection{Spatial Parameterization}
We begin by exploring parameterizations which efficiently map positions on our mesh to coefficients that encode appearance.

\tightboldparagraph{UV mapping} UV texture maps are the most ubiquitous representation for appearance. However, we found that current UV mapping tools cannot deal well with the complexity of our input mesh, which contains a lot of fine geometric detail (though concurrent work such as \citet{Nuvo} may provide a viable path).

\tightboldparagraph{Vertex Attributes} Prior mesh-based view synthesis methods like \citet{BakedSDF} store appearance coefficients at vertex attributes on the mesh and then interpolatie them across each face. Unfortunately, this requires the vertex density to be higher than the desired texture density, which results in prohibitively large and expensive meshes. This is not the case for us, as our meshes are drastically simplified, leading to large triangles in geometrically simple regions.

\tightboldparagraph{Volume Textures} We can directly associate a color value with each 3D position using a 3D volume texture.
A simple way to encode a volume sparsely is to subdivide the volume into blocks of $\DataBlockSize^3$ voxels and store only the nonempty blocks~\cite{SNeRG}. The choice of the block size $\DataBlockSize$ involves a trade-off: A small block size yields high compactness, but results in poor data locality, which leads to slow rendering. A large block size comes with high memory consumption, since any block that contains a single surface-adjacent voxel must be allocated. This also holds for alternative sparse data structures such as octrees~\cite{OctreeTextures} or spatial hashing~\cite{PerfectSpatialHashing}, since they equally depend on blocking for fast access.

\tightboldparagraph{Triplanes and Low-resolution Voxel Grid} Recently, it has been shown that volume textures can be encoded compactly with a combination of triplanes and a low-resolution voxel grid~\cite{MERF}. Both the triplanes and the low-resolution voxel grid are cache-friendly, leading to fast random access.

\begin{figure}[t!]
\centering
\includegraphics[width=\linewidth]{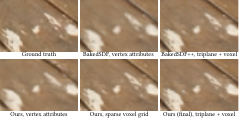}
\vspace{-0.3in}
\caption{\textbf{Comparison between different representations for mesh appearance.}
Replacing vertex attributes with a grid representation leads to sharper textures. There is almost no difference between ``voxel grid'' and the cheaper alternative ``triplane + voxel''.}
\label{fig:appearance_models}
\end{figure}

Table~\ref{tab:appearance_representations} shows that volume textures yield the highest quality, followed by the combination of triplanes and low-resolution voxel grid. The gap to vertex attributes is more pronounced, which you can also see in Figure~\ref{fig:appearance_models}: vertex attributes look blurry in geometrically simple regions.
Finally, while the triplane and grid combination uses less memory than and is much faster than volume textures, these representations look nearly identical in Figure~\ref{fig:appearance_models}.


\begin{table}[b]
\centering
\caption{Comparison between view-dependency encodings on \texttt{gardenvase} using our combination of triplanes and low-resolution voxel grid.} \vspace{-8pt}
\begin{tabular}{@{}l|cccc@{}}
& PSNR $\uparrow$ & SSIM $\uparrow$ & LPIPS $\downarrow$ & bytes $\downarrow$ \\ \hline
Spherical Gaussians & \textbf{26.02} & \textbf{0.807} & \textbf{0.157} & 24 \\
Spherical Harmonics & 25.65 & 0.797 & 0.166 & 27 \\
8-dim. Neural Feature & 25.18 & 0.781 & 0.179 & \textbf{8} \\
24-dim. Neural Feature & 25.72 & 0.798 & 0.164 & 24
\end{tabular}
\label{tab:view_dependent_color_encodings}
\end{table}

\subsection{View-Dependence}
We also investigate several encodings for view-dependent color:
\textbf{spherical harmonics} and \textbf{spherical Gaussians}, which are established formats for view-dependent colors in real-time view synthesis systems~\cite{PlenOctree, BakedSDF, Plenoxels}, and \textbf{neural feature vectors} that get decoded to a view-dependent color with a small MLP~\cite{SNeRG}.
As can be seen in Table~\ref{tab:view_dependent_color_encodings}, spherical Gaussians deliver the highest quality, while only requiring 24 bytes instead of 27 bytes per texel compared to spherical harmonics. To fairly compare neural feature vectors with spherical Gaussians, we choose a 24-dimensional neural feature vector to match memory consumption and bandwidth requirements. Our decoder uses the same architecture as other recent real-time view synthesis systems~\cite{SNeRG, MobileNeRF, MERF, SMERF}. Even with a large neural feature vector, this underperforms the non-neural baselines. As such, in our model, we use spherical Gaussians with the triplane and low-resolution grid combination, which gives the best trade-off between rendering speed, quality and memory consumption.

\begin{figure*}[t!]
\centering
\includegraphics[width=\linewidth]{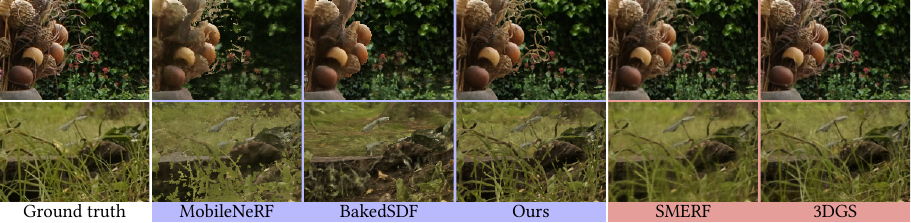}
\vspace{-0.25in}
\caption{Our method narrows the quality gap between \textbf{\textcolor{RowColorBIntense}{surface-based}} and  \textbf{\textcolor{RowColorAIntense}{volume-based}} methods when it comes to the reconstruction of thin structures.}
\label{fig:comparison_to_volume}
\end{figure*}

\subsection{Real-Time Implementation}
We implement a prototype web viewer for our representation based on Three.js. Since our meshes contain many tiny structures, anti-aliasing (AA) is critical during rendering. To avoid the computational expense of supersampling anti-aliasing (SSAA), we implement temporal anti-aliasing (TAA)~\cite{yang2020survey}, which amortises sampling across time and only needs a single sample per pixel per frame, as described in Appendix D. We find that the quality of our TAA is on par with the significantly more expensive SSAA, even when capturing frames under motion.

\section{Experiments}
We conduct experiments on the indoor and outdoor scenes of the challenging dataset from Mip-NeRF 360~\cite{Mip-NeRF_360}, where we compare our method in terms of quality, rendering speed, memory consumption, and storage impact to a range of volume-based and surface-based alternatives. We additionally conduct an ablation study to illustrate how our individual components contribute towards accurately reconstructing thin structures.

\subsection{Test-time Anti-Aliasing}
Using anti-aliasing for test-time rendering is crucial for high quality, as evidenced by Table~\ref{tab:anti_aliasing}. To this end, we study in our setting whether SSAA can be replaced by cheaper TAA without negatively affecting view synthesis quality. For this comparison, we always employ our previously-detailed $16\times$ supersampling strategy during training and only vary the anti-aliasing algorithm used for test-time rendering. Since TAA may introduce blur under motion, we also measure the quality of images that were captured after having moved the camera over a fixed number of frames to the target pose. As Table~\ref{tab:anti_aliasing} shows, TAA reaches comparable quality with the more expensive SSAA even when the camera is moving, making it a viable choice for our viewer application.

\subsection{Comparison with BakedSDF}
We compare with BakedSDF --- the state-of-the-art for real-time, mesh-based view synthesis --- in all scenes from Mip-NeRF 360~\cite{Mip-NeRF_360}. To disentangle whether differences in rendering quality between BakedSDF and our method stem from geometry or appearance, we also fit our best-performing appearance model (see Section~\ref{sec:view_dep_app}) to the meshes from BakedSDF, i.e.,
we encode appearance using a combination of triplanes and low-resolution voxel grid instead of vertex attributes. For a fair comparison, we also use the same supersampling strategy  for both BakedSDF and our method. We call this improved version BakedSDF++.

As can be seen in Figure~\ref{fig:appearance_models}, fitting our own appearance model to the meshes from BakedSDF leads to sharper textures, since with the combination of triplanes and low-resolution voxel grid, texture resolution is not bounded by vertex density. This indicates that this representation might be a viable alternative to vertex attributes even for dense meshes as produced by BakedSDF.

\begin{figure*}[h]
\centering
\includegraphics[width=\linewidth]{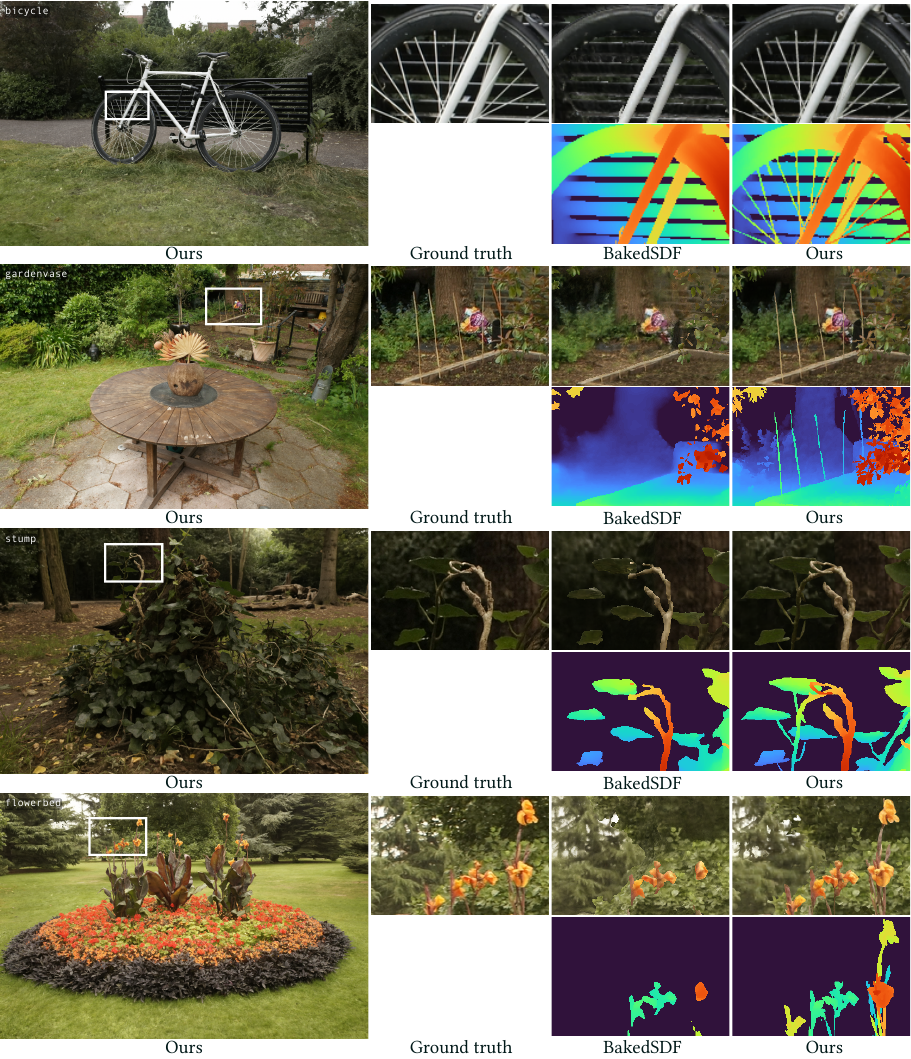}
\vspace{-0.25in}
\caption{Our method retains more geometric detail than BakedSDF. Bottom rows are visualizations of depth maps, for which we do not have ground-truth.}
\label{fig:bakedsdf_vs_ours}
\end{figure*}

More importantly, as shown in Figure~\ref{fig:bakedsdf_vs_ours}, our method is significantly better than BakedSDF at reconstructing thin structures, which are often absent for this baseline. Furthermore, Table~\ref{tab:detailed_comparision_to_bakedsdf} shows that our method outperforms BakedSDF++ in all key metrics, highlighting that our meshes are better suited for view synthesis than the meshes from BakedSDF.

\begin{table}[!]
\centering
\caption{
Quantitative results of our model on the outdoor and indoor scenes from mip-NeRF 360~\cite{Mip-NeRF_360}, with evaluation split for \textbf{\textcolor{RowColorAIntense}{volume-based}} and  \textbf{\textcolor{RowColorBIntense}{surface-based}} methods. Metrics not provided by a baseline are denoted with ``--''. The best metric within each family is shown in bold.} \vspace{-8pt}
\resizebox{\linewidth}{!}{
\Huge
\begin{tabular}{l|ccc|ccc}
& \multicolumn{3}{c|}{Outdoor Scenes} & \multicolumn{3}{c}{Indoor Scenes} \\
 & PSNR $\uparrow$ & SSIM $\uparrow$ & LPIPS $\downarrow$ & PSNR $\uparrow$ & SSIM $\uparrow$ & LPIPS $\downarrow$ \\ \hline
\rowcolor{RowColorA}  Instant-NGP~\shortcite{INGP} & 22.90 & 0.566 & 0.371 & 29.15 & 0.880 & 0.216 \\
\rowcolor{RowColorA}  MERF~\shortcite{MERF} & 23.19 & 0.616 & 0.343 & 27.80 & 0.855 & 0.271 \\
\rowcolor{RowColorA}  3DGS~\shortcite{3DGS} & 24.64 & 0.731 & 0.234 & 30.41 & 0.920 & 0.189 \\
\rowcolor{RowColorA}  Zip-NeRF~\shortcite{Zip-NeRF} & \textbf{25.68} & \textbf{0.761} & \textbf{0.208} & \textbf{32.65} & \textbf{0.929} & \textbf{0.168} \\
\rowcolor{RowColorA}  Shells~\shortcite{AdaptiveShells} & 23.17 & 0.606 & 0.389 & 29.19 & 0.872 & 0.285 \\
\rowcolor{RowColorA}  SMERF~\shortcite{SMERF} & 25.32 & 0.739 & 0.232 & 31.32 & 0.917 & 0.186 \\
\hline
\rowcolor{RowColorB}  Mobile-NeRF~\shortcite{MobileNeRF} & 21.95 & 0.470 & 0.470 & -- & -- & -- \\
\rowcolor{RowColorB}  BakedSDF~\shortcite{BakedSDF} & 22.47 & 0.585 & 0.349 & 27.06 & 0.836 & 0.258 \\
\rowcolor{RowColorB}  Ours (SSAA) & \textbf{23.94} & \textbf{0.680} & \textbf{0.263} & \textbf{27.71} & \textbf{0.873} & \textbf{0.227}
\end{tabular}
}
\label{tab:alldoor}
\end{table}

\begin{table}[t]
\centering
\caption{Rendering speed comparison in frames per second. Our method is significantly faster than \textbf{\textcolor{RowColorAIntense}{volume-based}} and \textbf{\textcolor{RowColorBIntense}{surface-based}} baselines and is the only method capable of real-time rendering on our test smartphone.} \vspace{-8pt}
\begin{tabular}{@{}l|cccc@{}}
Device & Smartphone & Laptop & Desktop \\
Resolution & $400 \times 750$ & $1280 \times 720$ & $1920 \times 1080$ \\\hline
\rowcolor{RowColorA} MERF~\shortcite{MERF} & 10 & 21 & 113 \\
\rowcolor{RowColorA} 3DGS~\shortcite{3DGS} & -- & -- & 176  \\ \hline
\rowcolor{RowColorB} BakedSDF~\shortcite{BakedSDF} & 19 & 81 & 412 \\
\rowcolor{RowColorB} Ours (TAA) & \textbf{67} & \textbf{448} & \textbf{927}
\end{tabular}
\label{tab:fps}
\end{table}

\begin{table}[t]
\centering
\caption{Comparison between test-time anti-aliasing algorithms on the outdoor scenes from the mip-NeRF 360 dataset~\cite{Mip-NeRF_360}. TAA achieves nearly the same fidelity as significantly more expensive SSAA.} \vspace{-8pt}
\begin{tabular}{@{}l|ccccc@{}}
 & PSNR $\uparrow$ & SSIM $\uparrow$ & LPIPS $\downarrow$ & FPS $\downarrow$\\ \hline
SSAA & 23.94 & \textbf{0.680} & \textbf{0.263} & 50 \\
TAA, stationary & \textbf{24.00} & \textbf{0.680} & 0.266 & 448 \\
TAA, under motion & 23.92 & 0.676 & 0.270 & 448 \\
No AA & 23.26 & 0.652 & 0.287 & \textbf{477}
\end{tabular}
\label{tab:anti_aliasing}
\end{table}

\begin{table}[t]
\centering
\caption{Comparison between BakedSDF, an improved version of BakedSDF (BakedSDF++) and our method on the outdoor scenes from mip-NeRF 360~\cite{Mip-NeRF_360}. Our method achieves higher view synthesis quality than BakedSDF++, which indicates that our compact meshes are better suited for view synthesis than BakedSDF's meshes.} \vspace{-8pt}
\begin{tabular}{@{}l|ccccc@{}}
& PSNR $\uparrow$ & SSIM $\uparrow$ & LPIPS $\downarrow$ & \#faces $\downarrow$ \\ \hline
BakedSDF & 22.47 & 0.585 & 0.349 & 40M \\
BakedSDF++ & 22.50 & 0.612 & 0.315 & 40M \\
Ours (SSAA) & \textbf{23.94} & \textbf{0.680} & \textbf{0.263} & \textbf{13M}
\end{tabular}
\label{tab:detailed_comparision_to_bakedsdf}
\end{table}

\begin{table}[t]
\centering
\caption{Quantitative results for geometric ablations on the outdoor scenes from the mip-NeRF 360 dataset~\cite{Mip-NeRF_360}.} \vspace{-8pt}
\begin{tabular}{@{}l|cccc@{}}
& PSNR $\uparrow$ & SSIM $\uparrow$ & LPIPS $\downarrow$ \\ \hline
(a) No supersampling & 23.38 & 0.645 & 0.292 \\
(b) No entropy loss & 23.21 & 0.635 & 0.293 \\
(c) $\Resolution = 2048$ instead of $\Resolution = 8192$ & 22.44 & 0.582 & 0.343 \\
Ours (SSAA) & \textbf{23.94} & \textbf{0.680} & \textbf{0.263}
\end{tabular}
\label{tab:ablations}
\end{table}

\subsection{Comparison with other Baselines}
We also compare our method with a broader set of baselines in terms of quality and rendering speed. We benchmark the volume-based baselines MERF and 3DGS and the surface-based method BakedSDF on a Google Pixel 8 Pro smartphone, a  MacBook M1 Pro (2022) laptop and a desktop equipped with an NVIDIA RTX 3090 graphics card. We report the harmonic mean of frames per second (FPS) on the outdoor scenes of the Mip-NeRF 360 dataset~\cite{Mip-NeRF_360}. In terms of rendering speed, our mesh-based representation outperforms all volume-based baselines, see Table~\ref{tab:fps}.
In terms of quality metrics, our method still lags behind the most recent volume-based baselines, as can be seen in Table~\ref{tab:alldoor}. However, the quality gap between surface-based and volume-based methods is significantly reduced, especially when it comes to the reconstruction of thin structures, as can be seen in Figure~\ref{fig:comparison_to_volume}.

\subsection{Geometry Ablations}
To investigate which elements contribute the most to geometric quality, we conduct an ablation study on the outdoor scenes from mip-NeRF 360~\cite{Mip-NeRF_360}. We focus here on the training of the initial opacity grid, which determines the quality of the mesh.

\begin{figure*}[h]
\centering
\includegraphics[width=\linewidth]{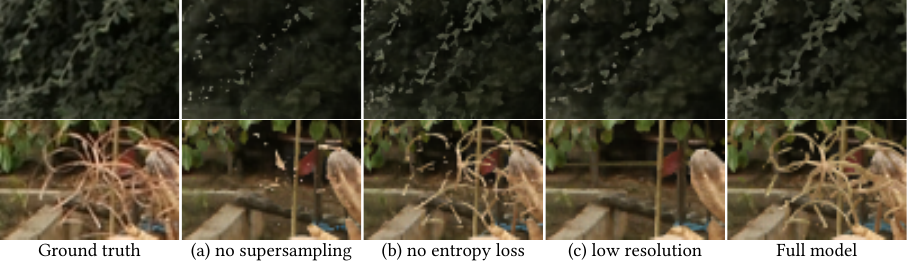}
\vspace{-0.25in}
\caption{\textbf{Qualitative results for geometric ablations.} 
Our model's ability to accurately reconstruct challenging thin structures depends critically on (a) casting multiple rays per pixel, (b) enforcing the entropy loss, and (c) employing a high resolution grid.}
\label{fig:ablations}
\end{figure*}

We train a variant of our model without supersampling (a). In this case, we only disable supersampling during training of the binary occupancy grid, but we still use supersampling for fitting the mesh appearance model and for computing quality metrics. This isolates the effect supersampling has on the quality of the obtained mesh. As shown by the top row of Figure~\ref{fig:ablations}, thin structures are hard to recover well without casting multiple rays per pixel during training.

Next, we train a variant of our model without the entropy loss (b). Since for this model, many opacity values do not become binary during the course of training, we define depth as the distance to the first voxel along the ray with an opacity value greater than $0.5$. Similar to (a), the effect of this is most pronounced for very thin structures such as the ones shown in Figure~\ref{fig:ablations}.

Finally, we decrease the resolution $\Resolution$ of the binary opacity grid (c). For this experiment, we only decrease the resolution of the initial binary opacity grid, but we use the same resolution of triplanes and the low-resolution voxel grid during mesh appearance fitting as for the full model. This isolates the effect geometric resolution has on mesh quality. As can be seen in Figure~\ref{fig:ablations}, a high resolution is crucial for reconstructing thin structures. Quantitative results for these ablations are given in Table~\ref{tab:ablations}.

\subsection{Storage Analysis}
Finally, we study how the individual components of our representation contribute to disk storage and memory consumption. We split our representation into a mesh and an appearance model. As we have shown experimentally, reconstructing thin structures requires a high grid resolution. As can be seen in Table~\ref{tab:vram_and_disk}, without any further processing, this leads to meshes with billions of faces, resulting in an impractical storage requirement of over 20 GiB. However, using mesh simplification and culling, the size of the mesh can be reduced by a factor of 100 to around 200 MiB. This results in the overall size of the representation being dominated by the appearance model, which occupies around $76\%$ of the overall storage.

\subsection{Limitations and Future Work}
Training-time supersampling adds a large computational overhead. The reconstruction of the underconstrained background of the scene is often highly noisy, which significantly increases the size of our meshes. This could potentially be mitigated with a smoothness regularizer. The concurrent work Nuvo presents a UV mapping method that is suited for high-detail meshes such as ours \cite{Nuvo}. Replacing our appearance representation with UV textures and obtaining more compact meshes using smoothness regularization could lead to further, significant speed-ups and memory savings.
Finally, we found the quality difference between our approach and volume-based methods larger in the indoor scenes. We attribute this to changes in illumination (e.g. shadows) between images that are difficult to capture on surfaces with a low-capacity view-dependence model.
Indeed, we found a large offline view-dependence network to yield significantly higher quality in these scenes.
The interpolated view-dependence networks from SMERF~\cite{SMERF} seem like a promising real-time alternative.

\section{Conclusion}
We have presented the first mesh-based view synthesis algorithm that is capable of reproducing subpixel structures in the input images by employing a high-resolution opacity grid combined with supersampling and a binary entropy loss. In contrast to volume-based alternatives, our method renders in real-time on affordable smartphones. Compared to BakedSDF, the previous state-of-the-art in mesh-based view synthesis, our method yields 3 times more compact meshes and achieves 1.46 dB higher PSNR in outdoor scenes.

%
%
%
%

\bibliographystyle{ACM-Reference-Format}
\bibliography{bibliography}

\begin{table}[b]
\centering
\caption{Memory consumption and storage impact of our method. (a) The dense mesh before simplification, (b) after simplification, (c) after culling. Simplification and culling are crucial for attaining practical mesh sizes. Results are averaged over all scenes from mipNeRF-360~\cite{Mip-NeRF_360}.} \vspace{-8pt}
\resizebox{\linewidth}{!}{
\begin{tabular}{@{}rl|cccc@{}}
& & (a) Dense Mesh & (b) + Simpl. & (c) + Culling\\ \hline
Mesh & \#vertices & 606M & 9M & \textbf{7M} \\
Mesh & \#faces & 1208M & 18M & \textbf{10M} \\
Mesh & VRAM & 20.28 GiB & 0.30 GiB &\textbf{0.19 GiB} \\
Mesh & DISK & 21.40 GiB & 0.32 GiB & \textbf{0.20 GiB} \\
\hline
Appearance & VRAM & 0.75 GiB & 0.75 GiB & \textbf{0.75 GiB} \\
Appearance & DISK & 0.65 GiB & 0.65 GiB & \textbf{0.65 GiB} \\
\hline \hline
Total & VRAM & 21.02 GiB & 1.05 GiB & \textbf{0.94 GiB} \\
Total & DISK & 22.05 GiB & 0.97 GiB & \textbf{0.85 GiB}
\end{tabular}
}
\label{tab:vram_and_disk}
\end{table}


\begin{appendix}

\section{Fusion Algorithm}
When converting our trained binary opacity grid into a triangle mesh, it is crucial to only instantiate geometry in regions that were sampled by the proposal MLP, since only these regions were supervised during training.
This can be achieved by rendering depth maps from the training viewpoints using the proposal MLP and creating surface voxels via unprojection. 
However, during training, some voxels may only be sampled in a fraction of the views and thus have incorrect opacity values. 
This leads to floating artifacts in the resulting mesh (see Figure~3 in the main paper).

We use volumetric fusion to filter these underconstrained voxels. 
Specifically, for each voxel, we count how for many depth maps the voxel is observed 1) in free space and 2) on the surface. 
As underconstrained voxels appear in very few training views, we can detect and discard them by only keeping voxels that are more frequently observed on the surface rather than in free space.

Another important motivation for volumetric fusion is that it outputs a dense implicit representation which can easily be converted into a hole-free mesh --- the preferred input for most simplification algorithms.
Consequently, we would like our fusion algorithm to also label unobserved voxels as ``inside'' or ``outside''.
A simple heuristic is to lo label unobserved voxels as ``inside''~\cite{VolumetricFusion}.
However, in rare cases, the proposal MLP does not sample the surface of an object, which carves large holes into the resulting mesh.
We address this by requiring several views to observe a voxel in free space before it can be labelled as outside.

This leads to our fusion algorithm. For each voxel, we count:
\begin{itemize}
    \item $\SurfaceCounter$: the number of views where it is observed on the surface.
    \item $\FreespaceCounter$: the number of views where it is observed in front of the surface, i.e. in free space.
    \item $\ObservedCounter$: the number of views in which it is observed at all.
\end{itemize}
Specifically, we project the voxel into each training view where we obtain a depth value $\SurfaceDepth$. 
Then, we increment $\SurfaceCounter$ if the voxel's depth is approximately the same as $\SurfaceDepth$. 
We increment $\FreespaceCounter$ if the voxels' depth is smaller than $\SurfaceDepth$. 
If the voxel is within the training camera's frustum, we increment $\ObservedCounter$.
Finally, we label the voxel as "inside" if any of the following conditions is met:
\begin{enumerate}
    \item $\SurfaceMultiplier\SurfaceCounter > \FreespaceCounter$
    \item $\FreespaceCounter < \FreespaceThreshold(\ObservedCounter)$
    \item $\SurfaceCounter = 0$ and $\FreespaceCounter = 0$
    \item $\ObservedCounter < 2$
\end{enumerate}

We multiply $\SurfaceCounter$ with a number $\SurfaceMultiplier > 1$ to bias the reconstruction towards surfaces, which helps preserve thin structures.
Weighing $\SurfaceCounter$ and $\FreespaceCounter$ equally leads to the erosion of objects, as voxels at object boundaries are not consistently sampled by the proposal MLP.
Rule (2) is motivated by the fact that sometimes the proposal MLP misses an entire object altogether.
In the unobserved interior of an object, $\SurfaceCounter$ will be equal to zero in this case, but since $\FreespaceCounter$ is nonzero, this leads to incorrectly labeling the voxel as outside.
Rule (2) fixes this by requiring a minimum number of views that need to observe a voxel to lie in free space for it to labelled as outside. This threshold needs to be dependent on then number of views that observe a voxel, since otherwise sparsely observed voxels are always labelled as inside.
Rule (3) ensures that completely unobserved parts of  the scene are labelled as inside.
The optional rule (4) ensures that only voxels are considered that are observed in at least two views, since parts of the scene that are only observed by a single camera are inherently underconstrained and thus only contribute geometric noise.
As can be seen in Figure~3 of the main paper, this algorithm effectively removes floating artifacts, while preserving thin structures.

\section{Scalable Mesh Conversion via Chunking}\label{sec:scalable_mesh_conversion}
To capture thin structures with a surface-based approach, a very high grid resolution of $8192^3$ is required. Processing such a high resolution grid in one pass requires too much memory. Fortunately, all of the steps (volumetric fusion, filtering, marching cubes, and simplification) in our pipeline can be executed in $1024^3$ chunks, which allows scaling to arbitrary resolutions. To avoid discontinuities at chunk boundaries, we configure the mesh simplification algorithm to keep boundary vertices intact. The final mesh can then be computed by concatenating sub-meshes and merging the duplicate boundary vertices. To speed up volumetric fusion, we only process voxels that are sufficiently close to an initial estimate of the surface. We obtain this initial estimate of the surface by unprojecting the depth values contained in the depth maps of all input views. We then quantize the resulting 3D points based on grid resolution ($8192^3$), which gives us a list of surface voxels. We then subdivide the scene into $16^3$ blocks and determine for each block whether it contains a voxel that is maximally $\VoxelDistance = 64$ voxels apart from a voxel in the previously computed list of observed surface voxels. During fusion we skip blocks that are not marked as alive. Since we are no longer densely computing the implicit representation, it is no longer guaranteed that running marching cubes results in a hole-free mesh, which is the preferred input for most simplification algorithm. However, we find that with our choice of $\VoxelDistance = 64$, only a moderate number of holes are introduced. These holes are usually not observed and therefore the quality of the reconstruction does not suffer. This technique also leads to slightly smaller meshes.

\section{Implementation Details and Hyperparameters}
\boldparagraph{Architecture} For the proposal MLPs and the MLP that predicts binary opacity values and view-dependent colors, we closely follow Zip-NeRF's \cite{Zip-NeRF} architecture based on a multi-resolution hash encoding \cite{INGP}. To bound opacity and color values between $0$ and $1$, we use a sigmoid activation function. Following \citet{MERF}, during mesh appearance fitting, we predict the values of the triplanes and low-resolution voxel grid with a hash grid-equipped MLP. For this MLP, we use the same architecture as the MLP that parameterizes the binary opacity grid. 

In unbounded scenes, regions that are only observed from far away can be represented with a low resolution. To achieve a resolution that smoothly decreases with the distance from the scene's center, we apply MERF's contraction function to each position $\mathbf{x}$ before querying the MLP:
\begin{equation} 
\operatorname{contract}(\mathbf{x}) = \begin{cases}
x &\text{if } \|\mathbf{x}\|_\infty \leq 1\\
\frac{x}{{\|\mathbf{x}\|_\infty}} &\text{if } x \neq \|\mathbf{x}\|_\infty > 1 \\
\left(2 - \frac{1}{|x|}\right) \frac{x}{|x|} & \text{if } x = \|\mathbf{x}\|_\infty > 1\end{cases}
\end{equation}
Before applying the contraction function, we scale input coordinates by a factor of $2.5$ to allocate more representation power to the foreground. For the standalone voxel grid, we use a resolution of $2048^3$. For the combination of triplane and low-resolution voxel grid, we use a resolution of $2048^2$ and $512^3$, respectively. 

\boldparagraph{Optimization}
For binary opacity grid optimization, we use Adam \cite{Adam} with an initial learning rate of $0.01$, a final learning rate of $0.001$ and 25K steps. For mesh appearance optimization, we use Adam with an initial learning rate of $0.0005$, a final learning rate of $0.00005$ and 100K steps. The learning rate is warmed up for $2500$ steps. For the binary entropy loss, we use a weight of $0.05$.

\boldparagraph{Simplification}
For mesh simplification, we need to specify a ratio $R$ that controls what fraction of original triangles should be kept. We want to simplify the background more aggressively than the foreground, since we find it to be less important for accurate view synthesis. As detailed in the previous section, simplification is executed on a chunk-by-chunk basis. The scene is subdivided into an $8^3$ grid of chunks. We define a chunk as lying in the background if its center lies outside of the $[-1,1]^3$ unit cube. For foreground chunks, we set $R$ to $0.03$. We simplify backgrounds chunks twice as aggressively by setting $R$ to $0.015$. In addition, we make sure that a background chunk contains at most 0.5M faces by adjusting $R$ accordingly.

\boldparagraph{Visibility Culling}
For visibility culling, we not only use the camera poses of the training images, but also generate $6$ additional poses for each training pose by adding random offsets and rotations to the original pose. Let $\mathbf{o}$ be the origin of the training camera let $\mathbf{d}$ be the direction the training camera faces to. We obtain a new origin $\mathbf{o}$ by applying isotropic Gaussian noise. To obtain a new direction $\hat{\mathbf{d}}$, we use the E3X library~\cite{RotationSampling} and draw a a uniform sample from an $\epsilon$-neighborhood of the the direction vector:
\begin{align}
    \hat{\mathbf{o}} &\sim \mathcal{N}(\mathbf{o}, \sigma^2 \mathbf{I}))\,, \\
    \hat{\mathbf{d}} &\sim \mathcal{U}(\{
        \mathbf{v} \in \mathbb{R}^3:
            \norm{\mathbf{v}-\mathbf{d}}_2 < \epsilon,
            \norm{\mathbf{v}}_2=1
    \})\,.
\end{align}
We find that the additional poses are crucial for avoiding visible holes in the final mesh.

\section{Temporal Anti-Aliasing}
We implement our temporal anti-aliasing strategy~\cite{yang2020survey} following industry best practices~\cite{karis2014taa}.
Namely, we jitter the projection matrix with a Halton$(2,3)$ sequence of length 16 and reproject the previous frame's color using the current depth buffer.
We then average the reprojected color with the current frame's color using an exponentially moving average with a blend factor of $0.05$.
To reduce blur from repeated resampling, we use a Lanczos kernel~\cite{duchon1979lanczos} with a radius of 3 for reprojection.
Finally, to limit ghosting artifacts for disoccluded content, we clip the reprojected color using variance-box~\cite{salvi2016taa} neighborhood clamping in the YCoCg color space~\cite{karis2014taa}.

Since TAA is known to cause blur under motion, we evaluate the quality of our test set images with a moving camera.
Given a target camera pose, we first extract its ``up'' vector $\mathbf{u}$ and its ``left'' vector $\mathbf{l}$. We then translate the camera from an initial position $\mathbf{p}+c(\mathbf{u}+\mathbf{l})$ to the target position $\mathbf{p}$ over a fixed number of frames $T = 100$: 
\begin{equation}
    \mathbf{p}(t) = \mathbf{p}+\lambda c(\mathbf{u}+\mathbf{l})
\end{equation}
where $\lambda = 1-\frac{t}{T-1}$ and the time step $t$ ranges from $0$ to $T-1$.
The factor $c = 0.05$ controls how far the initial position is from the target position. The camera's rotation is kept fixed during the entire trajectory. The frame used for computing quality metrics is captured when the camera has arrived at the target position.

\section{Frame Rate Benchmarking}
We follow the evaluation protocol from \citet{SMERF} and measure the average frame rate over the scene's test set camera poses. Following \citet{SMERF}, we render for each camera pose 100 frames and compute the average frame time. Similar to SMERF, for the browser-based viewer applications (BakedSDF \cite{BakedSDF}, MERF \cite{MERF} and our method), we measure frame rates that exceed the browser's frame rate limit by drawing each frame $k$ times before scheduling it for display. Following SMERF, we measure frame times with three different values of $k$, where we choose the initial value for $k$ to ensure that frame rate lies below 60 FPS. We then perform two additional measurements with larger values for $k$. Finally, for each frame, we report the minimum average frame time over $k$.

\end{appendix}

\end{document}